\documentclass[sigconf]{acmart}

\usepackage{booktabs} 

\usepackage{amsmath}
\usepackage{amssymb}
\usepackage{algorithm}
\usepackage{algorithmic}
\usepackage{graphicx}
\usepackage{float} 
\usepackage{latexsym}
\usepackage{array}
\usepackage{subfigure}
\usepackage{multirow}
\usepackage{makecell}
\usepackage{colortbl}
\usepackage{arydshln}
\usepackage{multicol}
\usepackage{courier}
\setcopyright{rightsretained}

\begin{document}
\title{Efficient Parallel Translating Embedding For Knowledge Graphs }

\copyrightyear{2017} 
\acmYear{2017} 
\setcopyright{acmcopyright}
\acmConference{WI '17}{August 23-26, 2017}{Leipzig, Germany}\acmPrice{15.00}\acmDOI{10.1145/3106426.3106447}
\acmISBN{978-1-4503-4951-2/17/08}

\author{Denghui Zhang, Manling Li, Yantao Jia, Yuanzhuo Wang, Xueqi Cheng}
\affiliation{%
  \institution{Institute of Computing Technology, Chinese Academy of Sciences}
  \city{No.6 Kexueyuan South Road, Zhongguancun, Haidian, Beijing} 
  \state{China} 
  \postcode{100190}
}
\email{zhangdenghui@ict.ac.cn,limanling@software.ict.ac.cn,jiayantao@ict.ac.cn,wangyuanzhuo@ict.ac.cn,cxq@ict.ac.cn}




\renewcommand{\shortauthors}{D. Zhang et al.}

\begin{abstract}
Knowledge graph embedding aims to embed entities and relations of knowledge graphs into low-dimensional vector spaces. 
Translating embedding methods regard relations as the translation from head entities to tail entities, which achieve the state-of-the-art results among knowledge graph embedding methods. 
However, a major limitation of these methods is the time consuming training process,  which may take several days or even weeks for large knowledge graphs, and result in great difficulty in practical applications. 
In this paper, we propose an efficient parallel framework for translating embedding methods, called ParTrans-X, which enables the methods to be paralleled without locks by utilizing the distinguished structures of knowledge graphs. 
Experiments on two datasets with three typical translating embedding methods, i.e., TransE \cite{bordes2013translating}, TransH \cite{Wang2014Knowledge}, and a more efficient variant TransE- AdaGrad \cite{Lin2015Learning} validate that ParTrans-X can speed up the training process by more than an order of magnitude.
\end{abstract}

%
%




\maketitle

\section{Introduction}

Knowledge graphs are 
structured graphs with various entities as nodes and relations as edges. 
They are usually in form of RDF-style triples $(h,r,t)$, where $h$ represents a head entity, $t$ a tail entity, and $r$ the relation between them. 
In the past decades, a quantity of large scale knowledge graphs have sprung up, e.g., Freebase \cite{bollacker2008freebase}, WordNet \cite{miller1995wordnet}, YAGO \cite{Mahdisoltani2014YAGO3}, OpenKN \cite{jia2014openkn}, and  have played a pivotal role in supporting many applications, such as link prediction, question answering, etc.
Although these knowledge graphs are very large, i.e., usually containing thousands of relation types, millions of entities and billions of triples, they are still far from complete. 
As a result, \emph{knowledge graph completion (KGC)} has been payed much attention to, which mainly aims to predict missing relations between entities under the supervision of existing triples. 

Recent years have witnessed great advances of translating embedding methods to tackle KGC problem. The methods represent entities and relations as the embedding vectors by regarding relations as translations from head entities to tail entities, such as TransE \cite{bordes2013translating}, TransH \cite{Wang2014Knowledge}, TransR\cite{Lin2015Learning}, etc.
However, the training procedure is time consuming, 
since they all employ stochastic gradient descent (SGD) to optimize a translation-based loss function, which may require days to converge for large knowledge graphs. 

\begin{table*}[h]
\centering
\caption{Complexity Analysis of Typical translating embedding Methods,  with  $d = 100$, $ep=1000$}
\newcommand{\tabincell}[2]{\begin{tabular}{@{}#1@{}}#2\end{tabular}}
\vspace{-8pt}
\begin{tabular}{|c|c|c|c|c|c|c|c|c|c|}
\hline &\multirow{2}{*}{ \tabincell{c}{Time \\Complexity}}&\multirow{2}{*}{\tabincell{c}{Model \\ Complexity}}&\multicolumn{2}{c|}{on FB15k}& \multicolumn{2}{c|}{on Freebase-rdf-latest }& \multicolumn{2}{c|}{on the whole Freebase}\\
\cline{4-9} &&& $T_{epoch}$ & $T_{total}$ &$T_{epoch}$  &$T_{total}$  & $T_{epoch}$  &$T_{total}$\\
\hline TransE& $\mathcal{O}(n_t \times d \times k)$  & $\mathcal{O}(dn_e +dn_{r})$ &4.5s&\tabincell{c}{4658s\\ $\approx$ 78 minutes} &\tabincell{c}{17,696s \\ $\approx$ 5 hours}&\tabincell{c}{18,323,395s \\ $\approx$ 212 days} &\tabincell{c}{29,781s \\ $\approx$ 8 hours} &\tabincell{c}{30,828,893s \\ $\approx$ 357 days}\\
\hline TransH& $\mathcal{O}(n_t \times d \times k)$ & $\mathcal{O}(dn_e +2dn_{r})$&6s&100 minutes &6.5 hours &273 days &11 hours & 459 days\\
\hline TransR& $\mathcal{O}(n_t \times d^2 \times k)$ &$\mathcal{O} (dn_e +dn_{r}+n_{r}d^{2})$ &473s&5 days &21.5 days &59 years &36 days &99 years\\
\hline
\end{tabular}
\label{complexityall}
\vspace{-6pt}
\end{table*}

For instance, \tablename \ref{complexityall}\footnote{The experiments are conducted on a dual Intel Xeon E5-2640 CPUs (10 cores each $\times$ 2 hyperthreading, running at 2.4 GHz) machine with 128GB of RAM. The kernel is Red Hat 4.4.7} shows the complexity of typical translating embedding methods,  
where $T_{total}$ stands for the total training time with $T_{epoch}$ for the time of each epoch, and one epoch is a single pass over all triples. 
$n_e$, $n_{r}$ and $n_t$ are the number of entities, relations and triples in  the knowledge graph respectively. $d$ is the embedding dimension which is the same for entities and relations in this case, and  $ep$ is the minimum epochs which used to be set to $1000$. It can be seen that the time complexity of TransE is proportional to $n_t$, $d$ and $ep$.
When $d$ is 100 and $ep$ is 1000, it will take 78 minutes for TransE to learn the embeddings of FB15k\footnote{https://everest.hds.utc.fr/lib/exe/fetch.php?media=en:fb15k.tgz}, which is a subset of Freebase with 483,142 training triples, and has been widely used as experimental dataset in knowledge graph embedding methods \cite{bordes2013translating,Wang2014Knowledge,Lin2015Learning,jia2016locally}. 
Nevertheless, Freebase-rdf-latest\footnote{http://commondatastorage.googleapis.com/freebase-public/ } is the latest public available data dump of Freebase with 1.9 billion triples, which results in approximately 3932 times the training time, namely, 212 days. 
Furthermore, the whole Freebase contains 
over 3 billion triples\footnote{https://github.com/nchah/freebase-triples, there are 3,197,653,841 triples in Freebase on May 2, 2016}, and 
it will take about 357 days to learn the embeddings of it.
Despite its large size, Freebase still suffers from data incomplete problem, 
e.g., 75\% persons do not have nationalities in Freebase \cite{Dong2014Knowledge}. 
On top of that, most improved variants of TransE employ more complex loss function to better train the embedding vectors, thus they possess higher time complexity or model complexity, and the training time of them will be even unbearable. For example, it will take more than 59 years for Freebase-rdf-latest 
when employing TransR, which is one of the typical improved variants and achieves far better performance than TransE.

 There have been attempts to resolve the efficiency issue of translating embedding methods for knowledge graphs. 
Pasquale\cite{minervini2015efficient} proposed TransE-AdaGrad 
 to speed up the training process by leveraging adaptive learning rates. 
 However, TransE-AdaGrad essentially reduces the number of epochs to converge, and still can not do well with large scale knowledge graphs.
In fact, with more and more computation resources available, 
it is natural and more effective to parallel these embedding methods, which will lead to significant improvement in training efficiency and can scale to quite large knowledge graphs if given sufficient hardware resources. 

However, 
it is challenging to parallel the translating embedding \\methods, since 
the training processes mainly employ stochastic gradient descent algorithm (SGD) or the variants of it. SGD is inherently sequential, as a dependence exists between each iteration.
Parallelizing translating embedding methods straightforwardly will result in collisions between different processors. 
For instance, an entity embedding vector is updated by two processors at the same time, and the gradients calculated by these processors are different. 
In this case, the diverse gradients are called \emph{collisions}. To avoid collisions, 
some methods \cite{Langford2009Slow} lock embedding vectors, which will slow the training process greatly as there are so many vectors. 
On the contrary, updating vectors without locks leads to high efficiency, but should be based on specific assumptions \cite{recht2011hogwild,Dean2012Large}. Since  the lock-free training process may result in poor convergence if adopting suboptimal strategy to resolve collisions. 

Our key observation of translating embedding methods is that  the update performed in one iteration of SGD is based on only one triple and its corrupted sample, which is not necessarily bound up with other embedding vectors. 
This gives us chance to learn the embedding vectors in parallel without being locked.
In this article, we analyze the distinguished data structure of knowledge graphs, and propose an efficient parallel framework for translating embedding methods, called ParTrans-X. 
It enables translating methods to update the embedding vectors efficiently in shared memory without locks. Thus the training process is greatly speeded up  with multi-processors, which can be more than an order of magnitude faster without lowering learning quality.

The contribution of this aritcle is:

1. We explore the law of collisions along with increasing number of processors, 
by modelling the training data of knowledge graph into hypergraphs.

2. We propose ParTrans-X framework to train translating methods efficiently in parallel. 
It utilizes the training data sparsity of large scale knowledge graphs, 
 and can be easily applied to many translating embedding methods. 


3. We apply ParTrans-X to typical translating embedding methods, i.e., TransE \cite{bordes2013translating}, TransH \cite{Wang2014Knowledge}, and a more efficient variant TransE-AdaGrad, and experiments validate the effectiveness of ParTrans-X on two widely used datasets. 

The paper is organized as follows. Related work is in Sec.2. The collision formulation is introduced in Sec.3 and ParTrans-X is proposed based on it in Sec.4.
Then, experiments demonstrate training efficiency of ParTrans-X in Sec.5, with conclusions in Sec.6. 

\section{Related Work}
In recent years, translating embedding methods have played a pivotal role in Knowledge Graph Completion, 
which usually employ stochastic gradient descent algorithm to optimize a translation-based loss function, i.e., 
\begin{equation}\label{eqlossall}
\vspace{-4pt}
L=\sum_{(h,r,t)} \sum_{(h',r,t')} \max\left[0,f_r(h,t)+M-f_r(h',t')\right],
\vspace{-4pt}
\end{equation} 
where $(h,r,t)$ represents the positive triple that exists in the knowledge graph, while $(h',r,t')$ stands for the negative triple that is not in the knowledge graph. 
$\max \left[0,\cdot \right]$ is the hinge loss 
, and $M$ is the margin between positive and negative triples. $f_r(h,t)$ is the score function to determine whether the triple $(h,r,t)$ should exist in the knowledge graph, which varies from different translating embedding methods.

A significant work is TransE \cite{bordes2013translating}, which heralds the start of translating embedding methods. It looks upon a triple $(h,r,t)$ as a translation from the head entity $h$ to the tail entity $t$, i.e., $\mathbf{h}+\mathbf{r} \approx \mathbf{t}$, and the score function is $f_r(h,t)=||\mathbf{h}+\mathbf{r}-\mathbf{t}||$, where $||\cdot||$ represents L1-similarity or L2-similarity. The boldface suggests the vectors in the embedding space, namely, $\mathbf{h},\mathbf{t} \in \mathbb{R}^{d}$, $\mathbf{r} \in \mathbb{R}^{d}$, 
where $d=d_{e}=d_{r}$ is the dimension of embedding space, $d_{e}$ the dimension for entities and $d_{r}$ for relations. 
Moreover, TransH \cite{Wang2014Knowledge} assumes that it is the projections of entities to a relation-specific hyperplane that satisfy the translation constraint, i.e., $f_r(h,t)=||\mathbf{h_\bot}+\mathbf{r}-\mathbf{t_\bot}||$, where $\mathbf{h_\bot}=\mathbf{h}-\mathbf{w_r^{\top}hw_{r}}$ and   $\mathbf{t_\bot}=\mathbf{t}-\mathbf{w_r^{\top}tw_{r}}$, with $\mathbf{w_r} \in \mathbb{R}^{d_{e}}$ as the normal vector of the hyperplane related to $r$.
Furthermore, TransR \cite{Lin2015Learning} employs rotation transformation to project the entities to a relation-specific space, i.e., $f_r(h,t)=||\mathbf{h_r}+\mathbf{r}-\mathbf{t_r}||$, where $\mathbf{h_r}=\mathbf{M_{r}h}$ and $\mathbf{t_r}=\mathbf{M_{r}t}$, and $\mathbf{M_{r}} \in \mathbb{R}^{d_{r} \times d_{e}}$ is the projection matrix relation to $r$.
Some works also 
involves more information to better embedding, e.g., 
paths \cite{Lin2015Modeling},  margins \cite{jia2016locally}.


Although this category of methods achieve the state-of-the-art results, the main limitation is the computationally expensive training process when facing large scale knowledge graphs
. Recently, a method TransE-AdaGrad \cite{minervini2015efficient} was proposed to reduce the training time of TransE by employing  AdaGrad \cite{duchi2011adaptive}, an variant of SGD, to adaptively modify the learning rate. 
Although the training time has been reduced 
greatly, there is still some way to go when facing large scale knowledge graphs.
With the computation resources greatly enriched, training in parallel seems to be a more reliable way to relieve this issue. Actually, there are some works, e.g., \cite{shaoparallel}, to parallel some graph computation paradigms, such as online query processing, offline analytics, etc.  Nevertheless, it is not easy to train translating embedding methods in parallel, since the main optimation algorithm SGD is born to run in sequence. 
The major obstacle to parallel SGD is the collisions between updates of different processors for the same parameter \cite{Ruder2016An},  to overcome which there are two main brunches of methods. 

The first brunch is to design a strategy to resolve collisions according to specific data structure. 
For example, Hogwild! \cite{recht2011hogwild} is a lock-free scheme works well for sparse data, which means that there is only a small part of parameters to update by each iteration of SGD. 
It has been proved that processors are unlikely to overwrite each other's progress, and the method can achieve a nearly optimal rate of convergence.
While the second brunch is to split the training data to to reduce collisions. 
Downpour SGD \cite{Dean2012Large}  
mainly employ DistBelief  \cite{Dean2012Large} framework, which divides the training data into a number of subsets, then the model replicas run independently on each of these subsets, and do not communicate with each other.
Inspired by this, TensorFlow \cite{Abadi2016TensorFlow} splits a computation graph into a subgraph for every worker and communication takes place using Send/Receive node pairs. 
Motivated by training large-scale convolutional neural networks for image classification, Elastic Averaging SGD (EASGD) \cite{Zhang2015Deep} 
reduces the amount of communication between local workers and the master to allow the parameters of local workers to fluctuate further from the center ones. 
There are also works to improve the performance in parallel settings, e.g.,  Delay-tolerant Algorithms for SGD \cite{Mcmahan2014Delay} adapts not only to the sequence of gradients, but also to the precise update delays that occur, inspired by AdaGrad.

However, these parallel framework are based on specific assumptions, 
and can not directly apply to translating embedding models without exploring distinguished data structures of knowledge graphs. 
 Therefore, we shall propose  a parallel framework for translating embedding models, called ParTrans-X, 
 as knowledge graphs are 
mainly in form of triples, and trained triple by triple, it will lead to particular parallel framework.

\section{Law of Collisions Emerging in KG}
As mentioned previously, there may exist collisions between processors  when they update the same embedding vector, which ends up being one of the most challenging aspects of parallelizing translating embedding methods. Hence, we explore the law of collisions emerging in this section.
At first we formulate the training data of knowledge graphs into hypergraphs. 
Then  the collisions in training process are further discussed based on this formulation.

\subsection{Hypergraph Formulation}

\begin{figure}[!h]
\centering
\scalebox{0.7}{\includegraphics{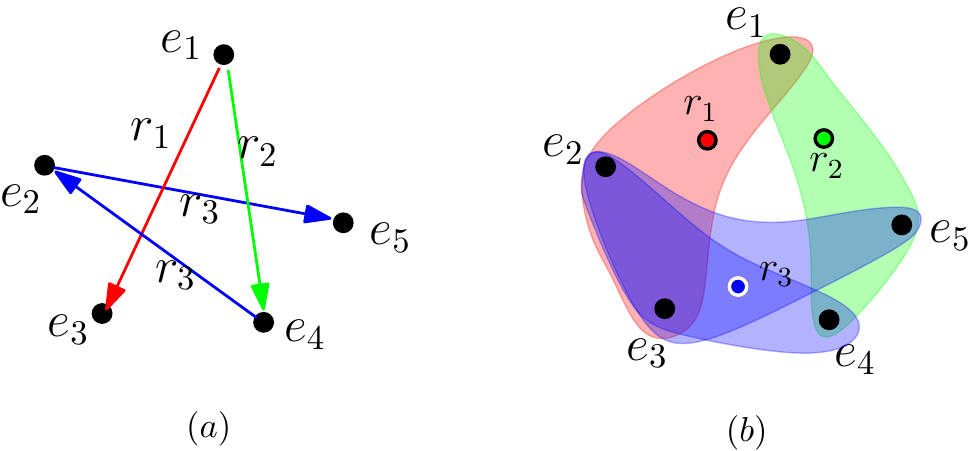}}
\vspace{-7pt}
\caption{A Knowledge Graph (a) and one of the Hypergraphs generated by its training data (b). }
\label{hypergraph}
\vspace{-7pt}
\end{figure}

Firstly, we model the knowledge graph formally as $G=(E,R,T)$, where $E$ is the set of entities with $R$ the set of relations, and $T$ is the set of triples $(h, r, t)$, in which $h,t \in E$ and $r \in R$. The  cardinalities of $E$, $R$ and $T$ are $n_e$, $n_r$ and $n$ respectively. In this graph, nodes are entities, and edges are triples that connecting nodes with a distinguished relation.
For example, the knowledge graph shown in \figurename \ref{hypergraph}(a), where black nodes stand for the entities in knowledge graphs and lines for relations, can be represented as $G=(E,R,T)$, where $E=\{e_1,e_2,e_3,e_4,e_5\}$, $R=\{r_1, r_1, r_3\}$ and $T=\{(e_1,r_2,e_4),(e_1,r_1,e_3),(e_2,r_3,e_5),(e_4,r_3,e_2)\}$. In this case, $n_e=5$, $n_r=3$ and $n=4$.

Secondly, the training data of knowledge graphs can be looked upon as hypergraphs. 
Recall the loss function of translating embedding methods in Eq.\eqref{eqlossall}, which means in one iteration of  SGD, only one positive triple $(h,r,t)$ and one negative triple $(h',r,t')$ are concerned. To be more clear, the data used in one iteration, i.e., $\left[(h,r,t), (h',r,t')\right]$, is called a \textit{sample}. 
Note that $(h',r,t')$ is constructed by substituting one entity $h' \in E$ or $t' \in E$ for $h$ or $t$ respectively, contributing to a corrupted triple $(h',r,t)$ or $(h,r,t')$, which is just simply denoted by $(h',r,t')$ following \cite{bordes2013translating}. Consequently, a sample corresponding to three entities, i.e., $h,t,h' or \ t'$, and one relation $r$. 
As a result, the training data can be formulated in to a 4-uniform hypergraph, in which all the hyperedges have  the same cardinality $4$. In this hypergraph, nodes are entities and relations, and edges are training samples containing 4 nodes, i.e., three entities and one relation. More formally, 
\begin{definition} 
The training data to embed the knowledge graph $G=(E,R,T)$ by translating embedding methods is organized as a 4-uniform hypergraph $H=(V,S)$, where $V = \{E \cup R\}$ is the set of entities or relations, and $S$ is the set of training samples $s$, where $s=\{h,r,t,h'(or \ t'): h,r,t,h',t' \in E, r \in R\}$.
\end{definition}

For example, the hypergraph in \figurename \ref{hypergraph}(b) is one of the hypergraphs generated by \figurename \ref{hypergraph}(a), where black nodes are entities and colored nodes are relations, and the colored blocks represent hyperedges. 
Here, different colors are related to different relations. 
For instance, for triple $(e_1, r_1, e_3)$, the negative triple sampled in \figurename \ref{hypergraph}(b)  is $(e_1,$ $ r_1,e_2)$, which contributes to a sample $s_1=\{e_1$, $r_1$,$e_2$,$e_3\}$, thus the hyperedge colored by red contains $e_1,e_2,e_3$ and $r_1$. Note that many other negative triples can be constucted, e.g., $(e_1, r_1,e_5)$ for triple $(e_1, r_1, e_3)$, and the hypergraph generated in  \figurename \ref{hypergraph}(b) is just an example. 
Similarly, the other samples in \figurename \ref{hypergraph}(b) are $s_2=\{e_1,r_2,e_4,e_5\}$, $s_3=\{e_2,r_3,e_3,e_5\}$ and $s_4=\{e_4,r_3,e_2,e_3\}$.

To better analyze the collisions between processors, we define the following statistics of the hypergraph  $H$. Given a hyperedge $s$,
\begin{align}
\vspace{-7pt}
\label{eqdeltae}
\sigma(s) = \{s': \exists r \in s \cap s', r \in R \}
\vspace{-7pt}
\end{align}
denotes the set of hyperedges containing the same relations with hyperedge $s$. 
\begin{align}
\vspace{-7pt}
\label{eqdeltaemax}
\hat{\sigma} = \max_{s \in S}|\sigma(s)|
\vspace{-7pt}
\end{align}
denotes the maximal number of  hyperedges containing same relations, where $|\cdot|$ denotes the cardinality.
\begin{align}
\label{eqdeltar}
\vspace{-7pt}
\rho (s) =  \{s': \exists e \in s \cap s', e \in E \} 
\vspace{-7pt}
\end{align}
denotes the set of hyperedges containing one or more same entities with hyperedge $s$. 
\begin{align}
\vspace{-7pt}
\label{eqdeltarmax}
\hat{\rho} := \max_{s \in S}|\rho(s)|.
\vspace{-7pt}
\end{align}
denotes the maximal number of  hyperedges containing same entities, where $|\cdot|$ denotes the cardinality the same as before.


\subsection{Collision Formulation}



In this section, we will verify that it is highly possible that few collisions happen when training by $p$ processors 
for large and sparse knowledge graphs. 
Let $X_{samp}$ represent the event that $p$ processors select $p$ different samples.
$X_{rel}$ represents the event that there are collisions between relations, i.e., different processors updates a same relation vector
, and $X_{ent}$ between entities similarly.
The verification is decomposed into two steps, 1) to prove it is quite likely that the processors handle different samples, i.e., $P(X_{samp}=1) \approx 1$, which is the prerequisite to no collisions; 2) to prove 
it is unlikely that these different samples correspond to the same relations or entities, i.e., $P(X_{rel}=0) \approx 1$ and  $P(X_{ent}=0) \approx 1$.

Supposing that for embedding methods and the knowledge graph, the training samples $ S=\{ s_1, s_2,\ldots, s_i,\ldots, s_{n} \}$
of size $n$ is drawn independent and identically distributed (i.e., i.i.d.) from some unknown distribution $\mathcal{D}$. 
Therefore, the probability of $s_i$ being selected $P_i$ is supposed to be 
\begin{equation}
\label{eqpi}
P_i=\frac{1}{n}.
\end{equation}
Moreover, according to i.i.d., it is reasonable to assume that the sample selecting process by $p$ processors is an observation from a Multinomial Distribution, i.e., selecting one sample from $n$ samples and repeated $p$ times. 
Let $x_i$ denote the number of processors that select $s_i$ during the same iteration of SGD, 
then the possibility of $s_i$ being selected by $c_1$ processors, $\ldots$,  $s_{n_e}$ being selected by $c_{n_e}$ processors is as follows,
\begin{align}
\label{eqmd}
\vspace{-7pt}
 P(x_1=c_1,..., & x_{n_e}=c_{n_e})= \nonumber\\
& \left\{
\begin{aligned}
\frac{p!}{c_1! c_2! \cdots c_{n_e}!}P_1^{c_1}...P_{n_e}^{c_{n_e}} &,  & \sum_{i=1}^{n_e}c_i=p \\
0 &,& otherwise
\end{aligned}
\right.
\vspace{-7pt}
\end{align}
where $\sum_{i=1}^{n_e}c_i=p$ indicates that there are $p$ and only $p$ samples being selected in the same iteration of SGD. 

\begin{theorem}
\label{theosamp}
For a knowledge graph with $n$ triples and training by $p$ processors in parallel, 
when $n$ is large and $p$ is relatively small, 
the possibility that $p$ processors select $p$ different samples is 
\begin{equation}
P(X_{samp}=1) \approx 1
\end{equation}
with probability at least $\gamma$, where 
\begin{equation}
\gamma=\prod \limits_{i=1}^{p-1}(1- \frac{i}{n}).
\end{equation}
\end{theorem}

\begin{proof}
Provided that samples selected by processors are different, it can be easily derived that $\forall s_i \in S, x_i \le 1.$
Then there are only $\binom {n} {p} $ sampling circumstances satisfying no collisions between samples, where  $p$ distinct samples are selected once, 
and other $n-p$ samples are not selected, e.g., $x_1=1,x_2=1,\cdots, x_{p}=1, x_{p+1}=0,\cdots,x_{n}=0$.
Therefore, according to Eq.\eqref{eqmd} and Eq.\eqref{eqpi}, 
\begin{align}
P(X_{samp}&=1) =\binom {n} {p}  \frac{p!}{\prod \limits_{i=1}^{p}1!}\prod \limits_{i=1}^{p} (P_{i})^1 
=\frac{n!} {(n-p)!} ( \frac{1}{n} )^{p}&\nonumber\\
&=\frac{n (n-1)(n-2) \cdots (n-p+1)}{{n}^{p}} &\nonumber
\end{align}

\begin{align}
&=(1-\frac{1}{n})(1-\frac{2}{n})\cdots(1-\frac{p-1}{n}) &\nonumber\\
&= \prod \limits_{i=1}^{p-1}(1-\frac{i}{n})&\nonumber
\end{align}

When $n$ is large and $p$ is relatively small, 
$P(X_{samp}=1) \approx 1$. 
\end{proof}

\begin{theorem}
\label{theorel}
For a knowledge graph with $n$ triples and training in $p$ processors in parallel, 
when $\frac{\hat{\sigma}}{n}$ is relatively small 
and $p < \frac{n}{\hat{\sigma}}+1$, 
we have the possibility of no relation in a collision  is 
\begin{equation}
P(X_{rel}=0) \approx 1
\end{equation}
with probability at least $\gamma$, where 
\begin{equation}
\gamma=\prod \limits_{i=1}^{p-1}(1- \frac{i\hat{\sigma}}{n}).
\end{equation}
\end{theorem}
\begin{proof}
Given that $p$ processors select $p$ different samples, the posibility of relations in a collision can be deduced according to conditional probability as follows,  
\begin{align}
\label{eqcondition}
P(X_{rel}=0) = P(X_{rel}=0|X_{samp}=1) \cdot P(X_{samp}=1),
\end{align}
where $P(X_{rel}=0|X_{samp}=1)$ is the possibility of  $p$  samples containing distinct relations being selected, which is supposed to be similar to sampling without replacement. 
More precisely, assuming a sample $s$ is selected randomly, then the next sample selected $s'$ should be from $S-\sigma(s)$, and the third sample $s''$ should be selected from in $S-\sigma(s) \cup \sigma(s')$. 
Accordingly, $P(X_{rel}=0|X_{samp}=1)$ is deduced as follows when $p < \frac{n}{\hat{\sigma}}+1$ is satisfied, 
\begin{align}
P(&X_{rel}=0|X_{samp}=1) &\nonumber\\
&= \sum_{s_1,s_2,...,s_{p} \in S} \frac{1}{n} \cdot \frac{|S-\sigma(s_1)|}{n-1} \cdot \frac{|S-\sigma(s_1) \cup \sigma(s_2)|}{n-2} \ldots &\nonumber\\
&  \frac{|S-\sigma(s_1) \cup \sigma(s_2) \cup \cdots \cup \sigma(s_{p-1})|}{n-(p-1)}&\nonumber\\
&\geq  \sum_{s_1,s_2,...,s_{p} \in S} \frac{1}{n} \cdot \frac{n-\hat{\sigma}}{n-1} \cdot \frac{n-2\hat{\sigma}}{n-2} \ldots   \frac{n-(p-1)\hat{\sigma}}{n-(p-1)}&\nonumber\\
&=(1-\frac{\hat{\sigma}-1}{n-1})(1-\frac{2(\hat{\sigma}-1)}{n-2})\cdots(1-\frac{(p-1)(\hat{\sigma}-1)}{n-(p-1)}) &\nonumber\\
&= \prod \limits_{i=1}^{p-1}(1- \frac{i(\hat{\sigma}-1)}{n-i})&\nonumber
\end{align}
By Eq.\eqref{eqcondition}, the possibility of no collisions between relations in different processors is 
\begin{equation}
P(X_{rel}=0)=  \prod \limits_{i=1}^{p-1}(1- \frac{i(\hat{\sigma}-1)}{n-i})\cdot \prod \limits_{i=1}^{p-1}(1- \frac{i}{n})=\prod \limits_{i=1}^{p-1}(1- \frac{i\hat{\sigma}}{n}).
\end{equation}
Note that $p>\frac{n}{\hat{\sigma}}+1$ results in $1-\frac{(p-1)(\hat{\sigma}-1)}{n-(p-1)} <0$, which means $\hat{\sigma}$ is so large that one or more processors will definitely select the same relation among $p$ processors, namely, $P(X_{rel}=0) =0$. 
Furthermore, when $\frac{\hat{\sigma}}{n}$ is relatively small
, $P(X_{rel}=0) \approx 1$.

\end{proof}

Similarly, the possibility of no entities in a collision can be derived as follows, and no more tautology here due to the limitation of length.
\begin{theorem}
\label{theoentity}
For a knowledge graph with $n$ triples and training in $p$ processors in parallel, 
when $\frac{\hat{\rho}}{n}$ is relatively small and $p < \frac{n}{\hat{\rho}}+1$, 
the possibility of no collisions between entities is 
\begin{equation}
P(X_{ent}=0) \approx 1
\end{equation}
with probability at least $\gamma$, where 
\begin{equation}
\gamma=\prod \limits_{i=1}^{p-1}(1- \frac{i\hat{\rho}}{n}).
\end{equation}
\end{theorem}

It is verified in Theorem\ref{theosamp}, Theorem\ref{theorel} and Theorem\ref{theoentity} that if $n$ is large and $\hat{\sigma}$ and $\hat{\rho}$ are relatively small, i.e., the knowledge graph is large and sparse, the number of processors $p$ can be very large with supportable collisions, which enables the training process to run in parallel. Motivated by this, we define \emph{sparsity} of training data in a knowledge graph by $\min(\frac{\hat{\sigma}}{n},\frac{\hat{\rho}}{n})$. The smaller its value is, the more processors can be used to parallel the training process. Actually, it is the large and sparse knowledge graphs that are in dire need of  parallel translating embedding methods. Since they are far from completion, but are too large to train in serial. Besides, since $\hat{\sigma}$ and $\hat{\rho}$ is deduced by the worst case, 
it is reasonable to assume that the average $\bar{{\sigma}}$ and $\bar{{\rho}}$  can better reflect the general structures in knowledge graphs, and the collisions will be less in practice. As a result, we suppose that it would still work well if the average $\bar{{\sigma}}$ and $\bar{{\rho}}$ are relatively small, as a few collisions will not affect the consistency. 
\subsection{Special Insights on Parallelizing TransE}
There is an interesting finding that TransE can be further parallelized than other translating embedding methods, since there are less collisions due to the distinguished score function $f_r(h,t)=||\mathbf{h}+\mathbf{r}-\mathbf{t}||$. More precisely, the gradient calculation of TransE when using $L2$-similarity is as follows, 
\begin{align}
\label{eqtransegrad}
& \mathbf{h}_k := \mathbf{h}_k-\eta \cdot 2(\mathbf{h}_k+\mathbf{r}_k-\mathbf{t}_k), \mathbf{h'}_k := \mathbf{h'}_k+\eta \cdot 2(\mathbf{h'}_k+\mathbf{r'}_k-\mathbf{t'}_k) \nonumber\\
&  \mathbf{r}_k := \mathbf{r}_k-\eta \cdot 2(\mathbf{h}_k+\mathbf{r}_k-\mathbf{t}_k) , \mathbf{r}_k := \mathbf{r}_k+\eta \cdot 2(\mathbf{h'}_k+\mathbf{r'}_k-\mathbf{t'}_k) \nonumber\\
&  \mathbf{t}_k := \mathbf{t}_k+\eta \cdot 2(\mathbf{h}_k+\mathbf{r}_k-\mathbf{t}_k) , \mathbf{t'}_k := \mathbf{t'}_k-\eta \cdot 2(\mathbf{h'}_k+\mathbf{r'}_k-\mathbf{t'}_k) 
\end{align}
where $ \mathbf{h}_k $ represents the $k$-th dimension of embedding vector $\mathbf{h}$, $k \in \{0,1,\cdots,d\}$, and $d$ is the dimension of embedding space. It can be seen that 
in TransE, the gradient of each dimension is independent of other dimensions, which means that the collisions between different dimensions of the same embedding vector will not disturb each other. That is to say, only the collisions between the  same dimension of the same embedding vector will matter in the training process of TransE.

\begin{figure}[htbp]
\centering
\scalebox{0.3}{\includegraphics{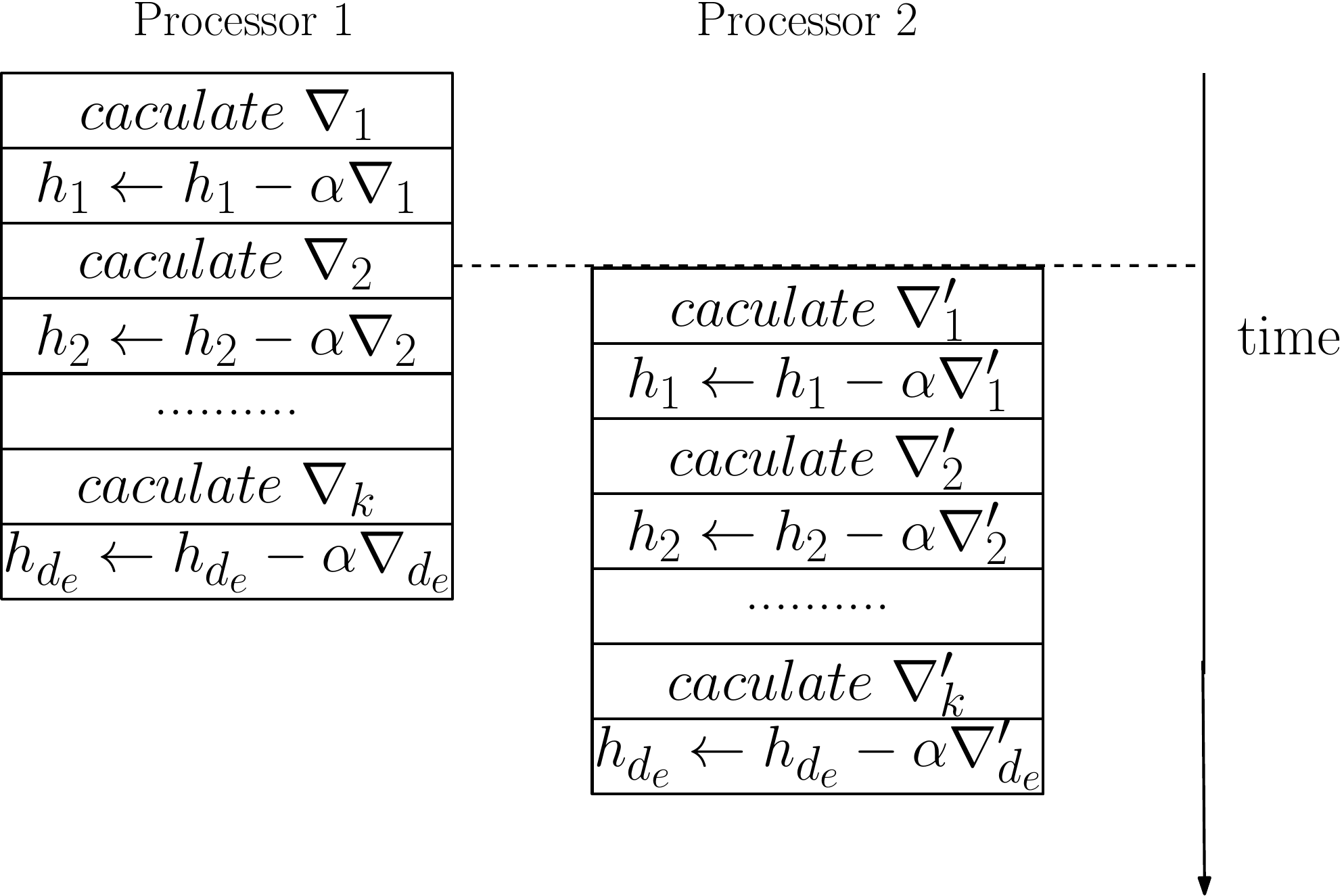}}
\vspace{-7pt}
\caption{Updating embedding vector $\mathbf{h}$ in parallel}
\label{multithread2}
\vspace{-7pt}
\end{figure}

For example, \figurename \ref{multithread2} shows the updating of $\mathbf{h}$ by two processors (Processor1 and Processor2) at the same time, where 
$\nabla_k$ is the gradient of $ \mathbf{h}_k$ calculated by Processor1, and $\nabla_k'$ by Processor2.
Normally, when Processor2 calculates the gradient $\nabla_k'$, the whole embedding vector $ \mathbf{h}$ will be involved, which  is half updated by Processor1. Obviously, this will result in training errors.   
On the contrary, if it is the training process of TransE in \figurename \ref{multithread2}, the calculation of $\nabla_k'$ by Processor2 only concerns the $k$-th dimension $\mathbf{h}_k$. As a result, there will no disturbance between Processor1 and Processor2, as long as the two processors are not performing update to the same dimension of the same embedding vector.

Consequently, the possibility of collisions emerging is greatly decreased for TransE. Since 
not only the entities or relations are the same one, but also the dimensions being updated are the same. 
Namely, the maximal degree of parallelism is far larger than other translating embedding methods.
This indicates that parallelizing without locks is ideally situated for TransE, and may scale well to extremely large knowledge graphs by given sufficient computation resources.

\section{The ParTrans-X Framework}

Inspired by the findings that collisions between processors are negligible when a knowledge graph is large and sparse, 
a parallel framework for these methods is designed, called ParTrans-X, and we will describe it in detail in this section. 

\subsection{Framework Description}

The pseudocode for implementation of ParTrans-X is shown in Algorithm 1.
As the embedding vectors are updated frequently, they are stored in shared memory and every processors can perform updates to them freely. 

The training process of ParTrans-X starts with initializing the embedding vectors according to Uniform or Bernoulli Distribution, where no parallel section is needed since it takes constant time.  
However we can parallel the learning process of each epoch, which is the most time consuming part. Running by $p$ processors in parallel can decrease the training epochs by $p$ times, i.e., the parallel training epoch is $ep' = \frac{ep}{p}$. To do this, we first determine the random sampling seed $seed[i]$ by calling $SEED\_RAND$ for the $i$-th processor. The random sampling seeds differ from each other 
to avoid same pseudo-random sequence for different processors.
Then, each processor performs embedding learning procedure epoch by epoch asynchronously (lines 5-12). One epoch is a loop over all triples. Each loop is done by firstly normalizing the entity embedding vectors following \cite{bordes2013translating}. Then a positive triple $P[i]^{(j)}=(h,r,t)$ is sampled from shared memory, where $i$ means that the current processor is $i$-th processor, and superscript $j$ stands for $j$-th epoch. According to $P[i]^{(j)}$, a negative triple $N[i]^{(j)}=(h',r,t')$ is generated by sampling a corrupted entity $h'$ (or $t'$) from shared memory, where $i$ and $j$ are the same as before. That is to say, a sample $S[i]^{(j)}$ is constructed by $P[i]^{(j)}$ and $N[i]^{(j)}$, which then be used to calculate the gradient $ \nabla[f_r(h,t)+M-f_r(h',t')]$ according to Eq.\eqref{eqlossall}, and update the embeddings of entities and relations $(h,r,t,h',t')^{(j+1)} \gets (h,r,t,h',t')^{(j)} -\eta\nabla[i]^{(j)}$.

\begin{algorithm}[htbp]
  \caption{ParTrans-X}
  \label{ParTrans-X}
  \begin{algorithmic}[1]
  \REQUIRE ~~\\
    Training triples $T=\{(h,r,t)\}$, entities and relations set $E$ and $R$, embedding dimension $d$, margin $M$, training epochs $ep$, the number of processors $p$;
  \ENSURE ~~ \\
     Embeddings of entities and relations;
  \STATE Initialize $r \in R$ and $e \in E$ by uniform distribution and persist them in the shared memory
  \FOR {$i \gets 0$ \TO $p$}   
    \STATE $ep' \gets \frac{ep}{p}$  \quad  \quad  \quad  \quad  \quad  \quad  \quad  \quad  \quad  \quad  \quad   \quad  \quad  \quad  $\triangleright$ In Parallel
    \STATE $seed[i] \gets SEED\_RAND(i)$ 
  \FOR {$j \gets 0$ \TO $ep'$}
    \LOOP
    \STATE $\mathbf{e} := \frac{\mathbf{e}}{||\mathbf{e}||}$ for each entity $e \in E$
    \STATE $P[i]^{(j)} \gets SAMPLE(T, seed[i])$
    \STATE $N[i]^{(j)} \gets SAMPLE\_NEG(P[i]^{(j)}, T, E, seed[i])$ 
        \STATE $S[i]^{(j)} \gets \left[P[i]^{(j)},N[i]^{(j)}\right]$
      \STATE $\nabla[i]^{(j)} \gets  \nabla[f_r(h,t)+M-f_r(h',t')],$ where $ h,r,t,h',t' \in S[i]^{(j)}$  \quad 
      \STATE $(h,r,t,h',t')^{(j+1)}\ \gets (h,r,t,h',t')^{(j)} -\eta\nabla[i]^{(j)}$ \quad 
  \ENDLOOP
  \ENDFOR
   \ENDFOR
    \STATE Generate embeddings of $E$ and $R$ after all processors finish
  \end{algorithmic}
\end{algorithm}

\subsection{Application to Typical translating embedding Methods}
The framework can be applied to many translating embedding methods,  which employ SGD or its variants to optimize the hinge loss 
with similar algorithm framework, and are only different in the score function $f_r(h,t)$ as mentioned in Sec.2, e.g., TransE, TransH and so on. Hence, the parallel algorithm of them can be obtained by applying the corresponding score function in Lines 11-12 of the pseudocode in Algorithm 1. 

For example, for \textbf{TransE}, the gradient updating procedure in Lines 11 is performed according to Eq.\eqref{eqtransegrad}. 
For \textbf{TransH}, which employs the score function $f_r(h,t)=||\mathbf{h_\bot}+\mathbf{r}-\mathbf{t_\bot}||$, the gradient updating procedure of $ \mathbf{h }$ in  Lines 11  is as follows,
\begin{align}
& \mathbf{h}_k := \mathbf{h }_k -\eta \cdot 2 \left[(\mathbf{h}_k -  \mathbf{w_r}^\top \mathbf{h} \mathbf{w_r}_k)+\mathbf{r}_k-(\mathbf{t}_k-  \mathbf{w_r}^\top \mathbf{t} \mathbf{w_r}_k) \right].
\end{align}
Namely,  ParTrans-X  has the flexibility to parallel many translating embedding methods, since they possess similar training process.

Moreover, ParTrans-X can be directly applied to the improved variant \textbf{TransE-AdaGrad}, since the training data sparsity of knowledge graph still holds. 
In one iteration of AdaGrad, it updates the embedding vectors according to the gradient from the previous iteration. 
Highly similar to SGD, AdaGrad can be easily parallelled using our framework by only performing a learning rate calculation procedure during the gradient update procedure, i.e., Line 12 of the pseudocode in Algorithm 1. 
For example, to parallel TransE-AdaGrad, the learning rate is determined adaptively by adding 
\begin{equation}
\eta^{(j)} := \frac{ \nabla^{(j)}}{\sqrt{\sum_{k=1}^{j} (\nabla^{(k)})^2}}{\eta*}
\end{equation} 
before Line 12 in Algorithm 1, where $j$ is the current epoch, with $\eta^{(j)}$ the learning rate of $j$-th epoch. $\nabla^{(k)},k<j$ represents all the previous gradient before $j$-th epoch. $\eta*$ is the initial learning rate.




\section{Experiment}
Firstly, 
we apply ParTrans-X to TransE, TransH and TransE-Adagrad in Sec.5.1. In Sec.5.2, experiment results demonstrate  excessive decline in training time by ParTrans-X,  
with scaling performance along with increasing number of processors shown in Sec.5.3.


\subsection{Experimental Settings}

The datasets employed are two representative datasets WN18 and FB15k, which are subsets of well-known knowledge graphs WordNet and Freebase respectively, and have been widely used by translating embedding methods \cite{bordes2013translating,Wang2014Knowledge,Lin2015Learning,jia2016locally}.  \tablename \ref{dataset} shows the statistics of them. Without loss of generality,  $\frac{\bar{\rho}}{n}$ and $\frac{\bar{\sigma}} {n}$ are also shown, and they are both small on WN18 and FB15k. Furthermore, it can be seen that the two datasets possess different characteristics. Namely, WN18 
possesses only 18 relations, which results in large possibility of collisions between relations. 
On the contrary, FB15k is 
less unbalanced in the number of entities and relations.

\begin{table}[htbp]
\centering
\caption{Two widely used datasets in KGs.}
\vspace{-7pt}
\label{dataset}
\begin{tabular}{|p{0.65cm}|p{0.65cm}|p{0.7cm}|p{0.78cm}|p{0.7cm}|p{0.7cm}|p{0.78cm}|p{0.78cm}|}
\hline
Data & \# Rel & \# Ent & \#Train & \#Valid & \#Test & $\bar{\sigma} /n$ & $\bar{\rho} / n$ \\
\hline
WN18 &  18 & 40,943 & 141,442 & 5,000 & 5,000 & 1.7e-1& 5.6e-4 \\
\hline
FB15k & 1,345 & 14,951 & 483,142 & 50,000 & 59,071 & 9.3e-3 & 2.3e-3\\
\hline
\end{tabular}
\vspace{-7pt}
\end{table}

\begin{table*}[htbp]
\newcommand{\tabincell}[2]{\begin{tabular}{@{}#1@{}}#2\end{tabular}}
\renewcommand{\arraystretch}{1.3}
\centering
\caption{Link prediction performance with all time measured in wall-clock seconds.}
\vspace{-7pt}
\begin{tabular}{|c|c|c|c|c|c|c|c|c|c|c|c|c|}
\hline
\multirow{3}{*}{Metric} & \multicolumn{6}{c}{WN18} & \multicolumn{6}{|c|}{FB15k} \\
\cline{2-13}
 & \multicolumn{2}{c|}{Mean Rank} & \multicolumn{2}{c|}{Hits@10} & \multirow{2}{*}{\tabincell{c}{Training \\ Time(s)}} & \multirow{2}{*}{\tabincell{c}{Speedup \\Ratio}} & \multicolumn{2}{c|}{Mean Rank} & \multicolumn{2}{c|}{Hits@10} & \multirow{2}{*}{\tabincell{c}{Training\\ Time(s)}} & \multirow{2}{*}{\tabincell{c}{Speedup \\Ratio}} \\
\cline{2-5} \cline{8-11} & Raw & Filter & Raw & Filter & &  &  Raw & Filter & Raw & Filter & & \\
\hline
TransE &214  &203  &58.2  &65.9 &473 & -&184  &73 & 44.5& 60.7& 4658& -\\
\textbf{ParTransE} & 217 &206 &55.7 &63.1 & \textbf{54}& \textbf{9} & 185 &74 & 44.4& 60.5& \textbf{364}&\textbf{13} \\
\hdashline[0.8pt/2pt]
TransE-AdaGrad & 209  & 197& 68.9 &77.7 &100 &4.7 &185 &69 &45.3 &62.3 &496 &9 \\
\textbf{ParTransE-AdaGrad} &  219 &208 &67.7 &76.2 & \textbf{17} &\textbf{28}{ (4.7$ \times \textbf{6}$)} &186 &70 &44.9 &61.9 &\textbf{42}&\textbf{111}{ (9$ \times \textbf{12}$)} \\
\hline
TransH & 227  &216 & 66.5&75.9  &637 &- & 183&60 &46.6 &65.5 &6066 &- \\
\textbf{ParTransH} &215 &203 & 66.8& 76.6 &\textbf{134} &\textbf{4.8} &183  &60 &46.8 &65.7 &\textbf{474} & \textbf{13}\\
\hline
\end{tabular}
\vspace{-7pt}
\label{relationresult}
\end{table*}

To tackle the $KGC$ problem, experiments are conducted on the link prediction task
which aims to predict the missing entities $h$ or $t$ for a triple $(h, r, t)$. Namely, it predicts $t$ given $(h, r)$ or predict $h$ given $(r, t)$. Similar to the setting in \cite{bordes2013translating}, the task returns a list of candidate entities from the knowledge graph. 

To evaluate the performance of link prediction, we adopt Mean Rank and Hits@10 under ``Raw'' and ``Filter'' settings as evaluation measure following  \cite{bordes2013translating}. Mean Rank is the average rank of the correct entities, and Hits@10 is proportion of correct entities ranked in top-10. It is clear that a good predictor has low mean rank and high Hits@10.  
This is called ``Raw'' setting, and ``Filter'' setting filters out the corrupted triples which are correct. 

To evaluate the speed up performance, we adopt Training Time and Speed-up Ratio as evaluation measures, where Training Time is measured using wall-clock seconds. 
Speed-up Ratio is 
\begin{equation}
Speedup \quad Ratio=\frac{t_{serial}}{t_{parallel}},
\end{equation} 
where $t_{serial}$ is the training time in serial, and $t_{parallel}$ is the training time under parallel methods.

Baselines include typical translating embedding methods, TransE, TransH and TransE-Adagrad, which can all be trained in parallel using the ParTrans-X framework, denoted by ParTransE, ParTransH and ParTransE-Adagrad respectively in \tablename \ref{relationresult}.
Note that TransE  and TransH adopt the programs publicly available\footnote{https://github.com/thunlp/KB2E}, which are the most efficient serial versions to our knowledge, and TransE-Adagrad is implemented based on TransE. 


Each experiment is conducted 10 times and the average is taken as results, with all time measured in wall-clock seconds. Our experiments are carried out on dual Intel Xeon E5-2640 CPUs, and each of them possesses 10 physical cores 20 logical cores and running at 2.4 GHz.
The machine has 128 GB RAM and runs Red Hat 4.4.7.
The language used  is C++ and the program is compiled with the gcc compiler version 6.3.0. We use OpenMP for multithreading, each thread binds a processor.








\subsection{Link Prediction Peformance of ParTrans-X}

Experiments on each baseline and its parallel implementation in ParTrans-X employ the same hyper-parameters, which are decided on the validation set. The learning rate $\eta$ during the stochastic gradient descent process is selected among \{0.1,0.01,0.001\}, the embedding dimension $d_e$ and $d_r$ are selected in \{20,50,100\}, the margin $M$ between positive and negative triples is selected among \{1,2,3,4\}. For TransE and ParTransE, the parameters are  $\eta=0.01, d_e=d_r=20, M = 3$ on WN18, and $\eta=0.001, d_e=d_r=100, M = 4$ on Fb15k. For TransH and ParTransH, the parameters are $\eta=0.01, d_e=d_r=20, M = 3$ on WN18, and $\eta=0.001, d_e=d_r=100, M = 3$ on Fb15k. For TransEAdaGrad and ParTransE-AdaGrad, the parameters are $\eta*=0.3, d_e=d_r=50, M = 4$ on WN18, and $\eta*=0.1, d_e=d_r=100, M = 3$ on Fb15k.
All the experiments employ $L1$-similarity. 
ParTransE, ParTransH and ParTransE-AdaGrad all run in $20$ processors for both datasets.

It can be  observed from \tablename \ref{relationresult}  that:

1.   Link prediction performance in parallel is as good as the serial counterparts on both WN18 and FB15k, which demonstrates that ParTrans-X will not affect embedding performance.

2. The training time is greatly reduced by ParTrans-X. On WN18, TransE-AdaGrad only speeds up TransE by 4.7 times, compared to our 28 times. 
On FB15k, the training time of TransE is reduced from more than 1 hour to less than 1 minute by ParTransE-AdaGrad.

3.  ParTrans-X achieves higher speedup ratio on FB15k than on WN18. 
Since FB15k has far more training triples than WN18, the time of each epoch on FB15k is much longer than WN18.  As a result, the overhead of multi-threading is less important compared to the whole training time on FB15k,  which leads to a higher speedup ratio.
It further validates the superiority of ParTrans-X to handle the data with large size.

4. ParTrans-X achieves enormous improvement on training time when applying to TransEAdaGrad, especially on FB15k, where the speedup ratio has been improve to 111 from 9. Since AdaGrad decreases the total epochs needed by making the convergence come earlier, and ParTrans-X reduce training time by running in parallel, the two different strategies can achieve higher speedup ratio when combined.



\begin{figure}[htbp]
\begin{minipage}[t]{0.45\linewidth}
\centering
\includegraphics[width=1.5in]{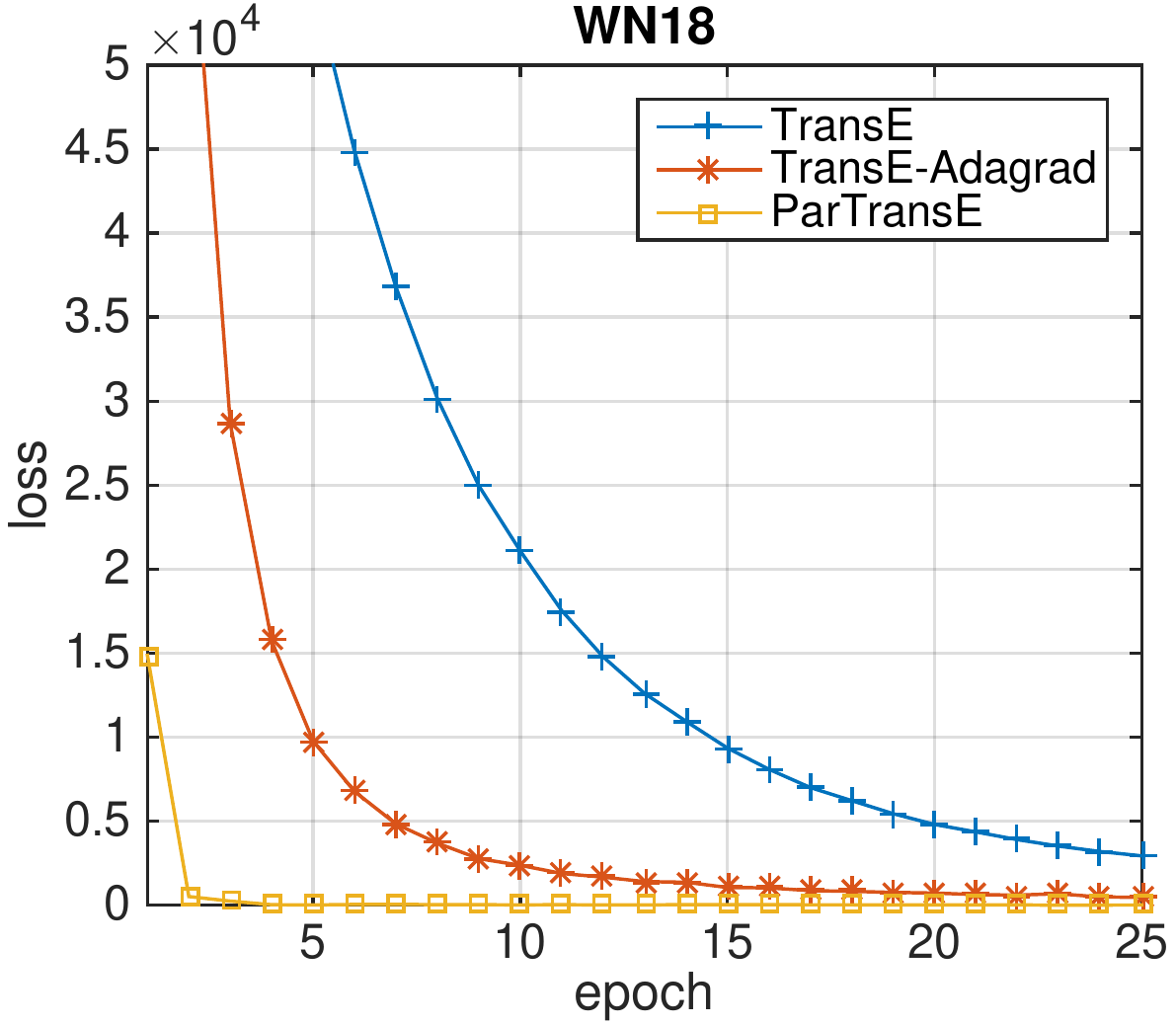}
\vspace{-20pt}
\caption*{WN18}
\vspace{-5pt}
\end{minipage}
\begin{minipage}[t]{0.45\linewidth}
\centering
\includegraphics[width=1.5in]{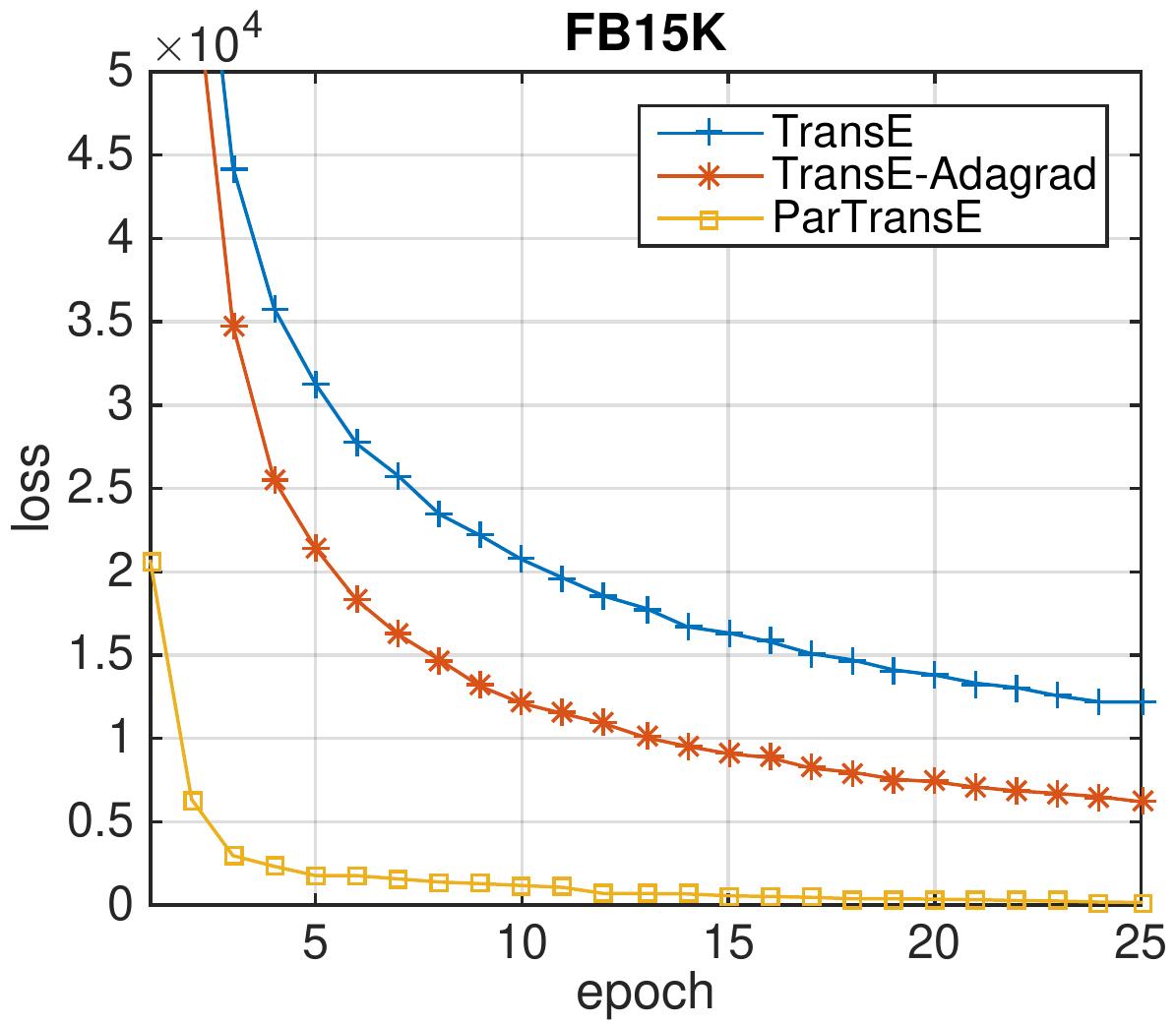}
\vspace{-20pt}
\caption*{FB15k}
\vspace{-5pt}
\end{minipage}
\caption{The descent process of loss.}
\label{losses1}
\vspace{-10pt}
\end{figure}

Moreover, the descent process of loss for the three algorithms on WN18  and  FB15k is shown in \figurename \ref{losses1}.  It can be seen that, for both datasets, the loss optimizing by ParTrans-X has already  fallen sharply  in the preceding epochs, and it yields sensibly lower values of the loss than TransE-AdaGrad and TransE even after a few iterations($<$ 5 epoches). 
 Still, ParTrans-X performs better on FB15k than WN18, shows that it is more effective on large data size.

\subsection{Scaling Results for Multi-Processors}


Furthermore, we carry out a number of experiments to test if the implementations scale with increasing number of processors. 
We mainly analyze two aspects of experiment results, i.e., the training time and the link prediction performance.

\begin{figure}[htbp]
\begin{minipage}[t]{0.45\linewidth}
\centering
\includegraphics[width=1.5in]{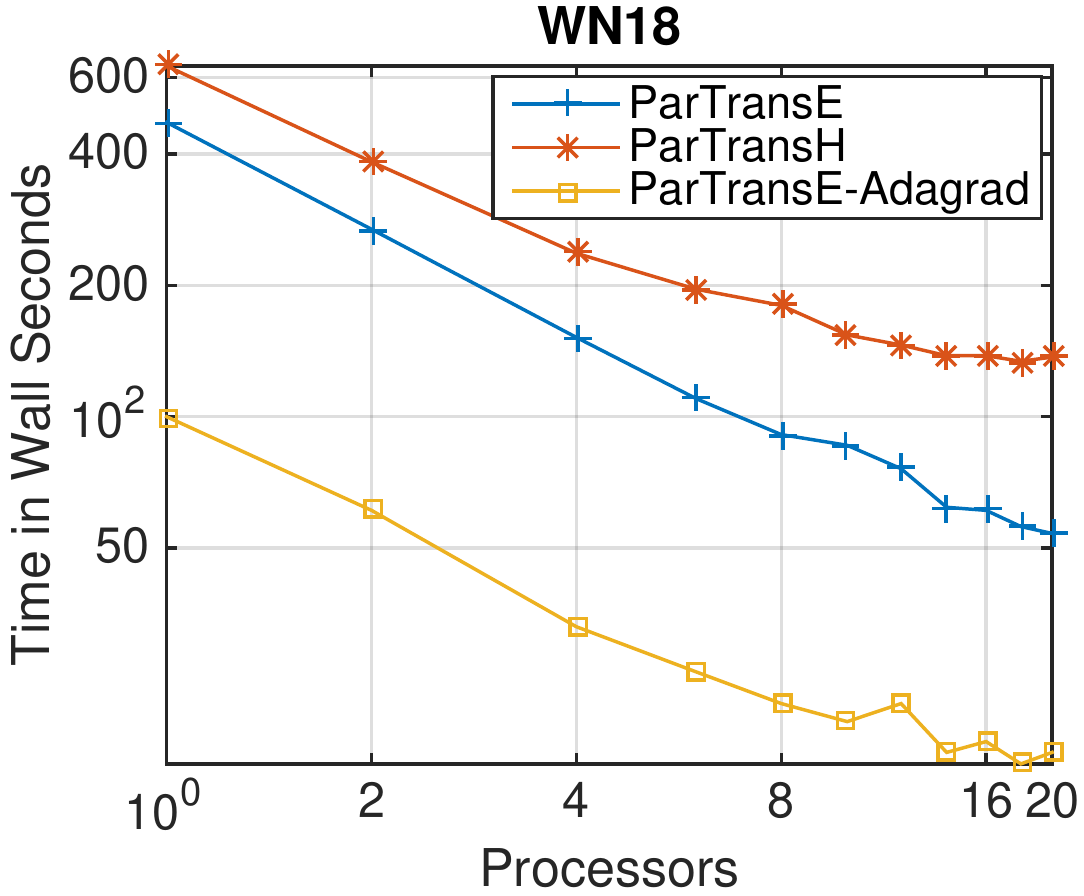}
\vspace{-15pt}
\caption*{WN18}
\end{minipage}
\begin{minipage}[t]{0.45\linewidth}
\centering
\includegraphics[width=1.5in]{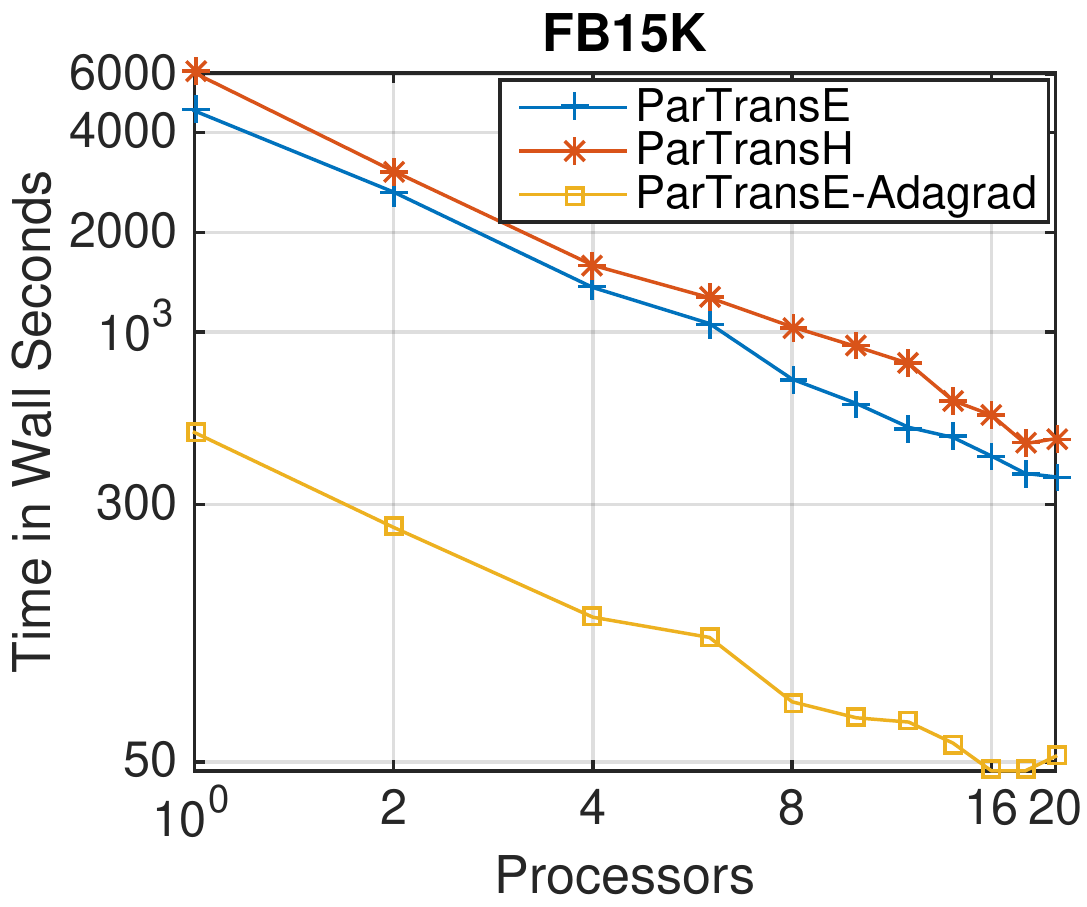}
\vspace{-15pt}
\caption*{FB15k}
\end{minipage}
\vspace{-7pt}
\caption{Log-log plot of Training Time along with number of processors}
\label{traintime}
\vspace{-10pt}
\end{figure}

\figurename \ref{traintime}  shows the log-log plot of the training time in wall-clock seconds for different number of processors. We can observe that the training time continue to decrease along with the increasing number of mutli-processors on both WN18 and FB15k. While the absolute training time of ParTransE-AdaGrad is better than ParTransE, which is better than ParTransH, consistent with the previous result. Moreover, the total training time of ParTransE-AdaGrad drops sharply when processor number is less than four, it is because the training time of ParTransE-AdaGrad with few processors is fairly short, the increase of communication time cost  with the more processors has larger effect on the total training time compared with other methods,  which leads to small decline.

\begin{figure}[htbp]
\begin{minipage}[t]{0.45\linewidth}
\centering
\includegraphics[width=1.5in]{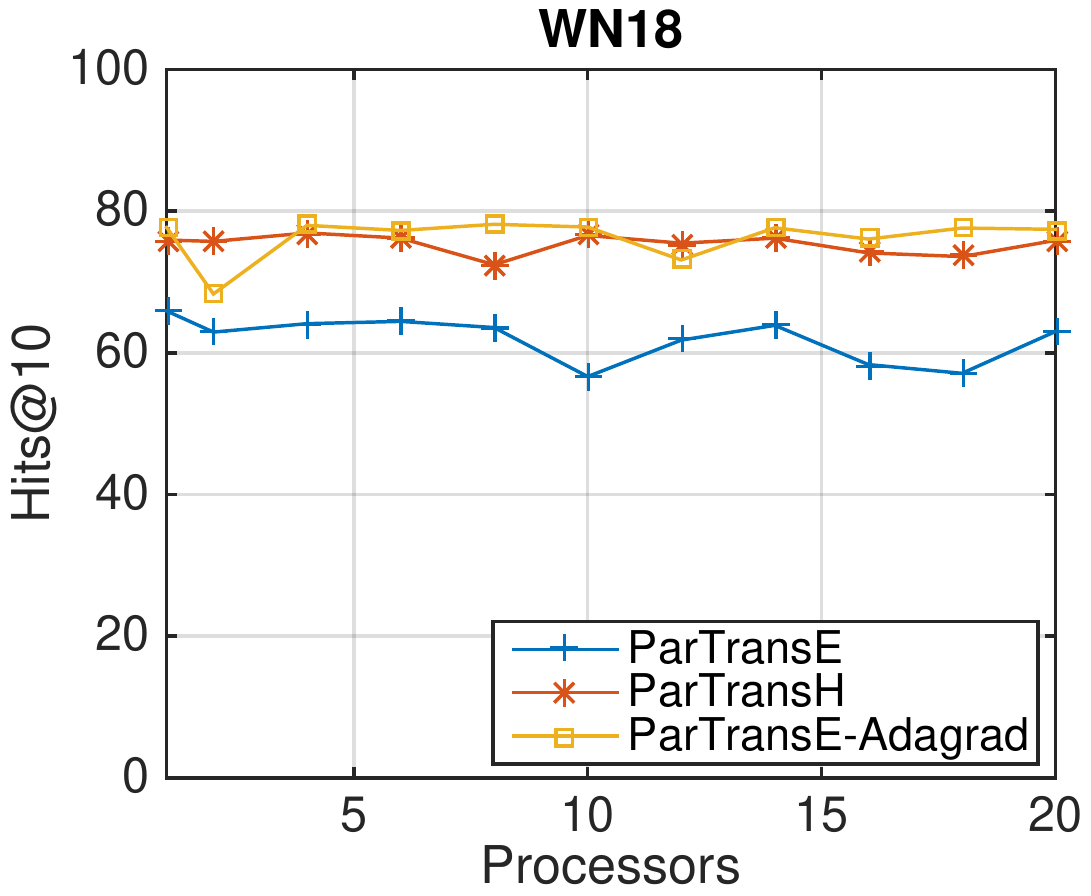}
\vspace{-15pt}
\caption*{WN18}
\end{minipage}
\begin{minipage}[t]{0.45\linewidth}
\centering
\includegraphics[width=1.50in]{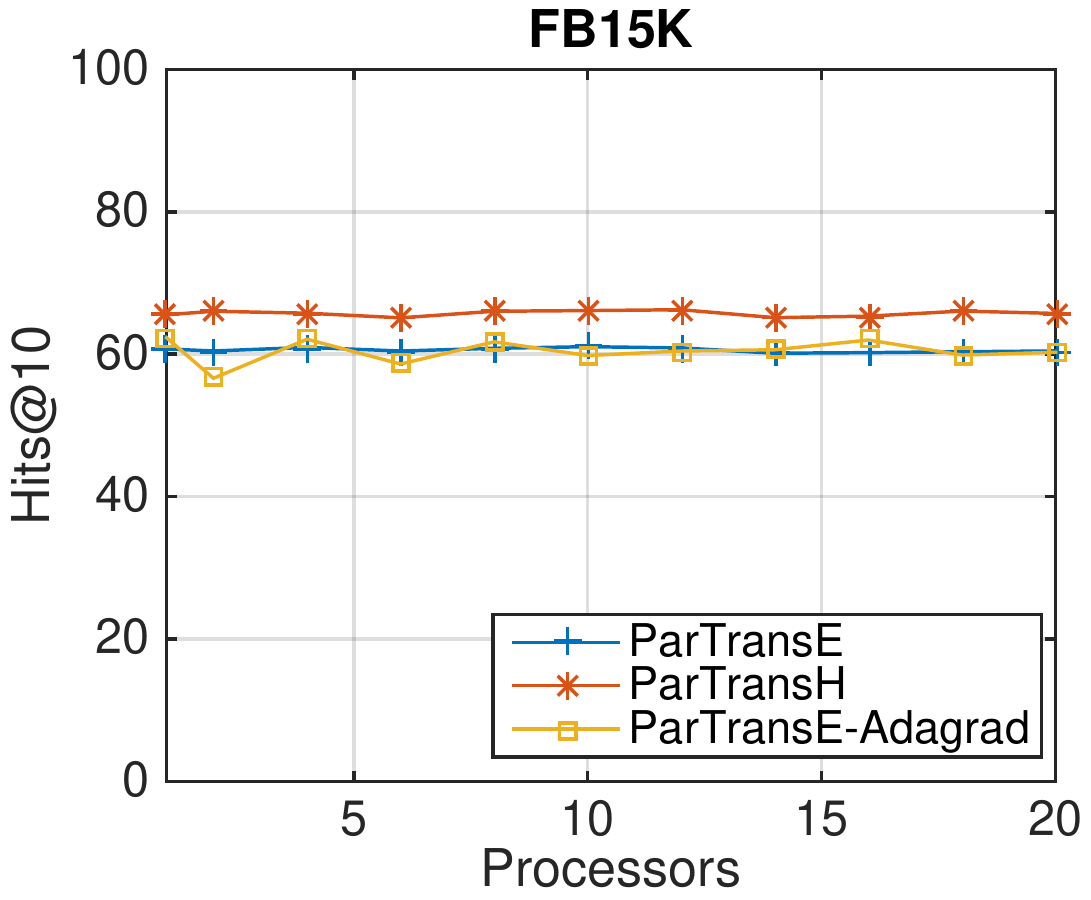}
\vspace{-15pt}
\caption*{FB15k}
\end{minipage}
\vspace{-7pt}
\caption{Hits@10 performance along with number of processors}
\label{performance}
\vspace{-10pt}
\end{figure}


 The predictive performance measured by Hits@10 along with increasing number of processors is shown in \figurename \ref{performance}. It can be seen that ParTransE, ParTransE-AdaGrad and ParTransH always maintain good performance, which  validates the applicability  and superiority of  ParTrans-X. Note that the performance on FB15k is more stable than WN18, since there are more training triples in FB15k, and the model will learn more sufficient so that the stability of predictive performance is better on FB15k, which validates the superiority of ParTrans-X on large data size.

\section{Conclusion}
In this paper, we explore the law of collisions emerging in knowledge graphs by modelling training data to hypergraphs. Our key observation is that 
one learning iteration only concerns few embeddings, which is not necessarily bound up with others, thus the probability of collisions between different processors can be negligible. 
Based on this assumption, we propose an efficient parallel framework for translating embedding methods, called ParTrans-X. It employs the intrinsic sparsity of training data in large knowledge graphs, which enables the embedding vectors to be learnt without locks and not inducing errors.  
Experiments validate that ParTrans-X can speed up the training process by more than an order of magnitude, without degrading embedding performance. The source code of this paper can be obtained from \href{https://github.com/zdh2292390/ParTrans-X}{here}\footnote{https://github.com/zdh2292390/ParTrans-X}.

\section{Acknowledge}
We thank Jun Xu and the anonymous reviewers for valuable suggestions. The work was funded by National Natural Science Foundation of China (No. 61572469, 61402442, 91646120,61572473, 61402\\022), the National Key R\&D Program of China (No. 2016QY02D0405, 2016YFB1000902), and National Grand Fundamental Research 973 Program of China (No. 2013CB329602, 2014CB340401).

\bibliographystyle{ACM-Reference-Format}
\bibliography{references}

\end{document}